\documentclass[11pt,a4paper]{article}
\usepackage[hyperref]{acl2021}
\usepackage{times}
\usepackage{latexsym}
\usepackage{graphicx}
\usepackage{amsmath}
\usepackage{multirow}
\usepackage{amssymb}

\usepackage{microtype}

\aclfinalcopy 

\title{Joint Event Extraction via Structural Semantic Matching}

\author{Haochen Li \\
  Peking University \\
  \texttt{Haochenli@pku.edu.cn } \\\And
  Tianhao Gao \thanks{these authors have contributed equally} \\
  Peking University \\
  \texttt{gaotianhao@pku.edu.cn} \\}
\date{}

\begin{document}

\maketitle
\begin{abstract}
Event Extraction (EE) is one of the essential tasks in information extraction, which aims to detect event mentions from text and find the corresponding argument roles. The EE task can be abstracted as a process of matching the semantic definitions and argument structures of event types with the target text. This paper encodes the semantic features of event types and makes structural matching with target text. Specifically, Semantic Type Embedding (STE) and Dynamic Structure Encoder (DSE) modules are proposed. Also, the Joint Structural Semantic Matching (JSSM) model is built to jointly perform event detection and argument extraction tasks through a bidirectional attention layer. The experimental results on the ACE2005 dataset indicate that our model achieves a significant performance improvement.
\end{abstract}

\section{Introduction}
An event is a specific occurrence involving participants, which is can frequently be described as a change of state\footnote{The ACE English event guidelines}. Event Extraction (EE) is one of the essential tasks in information extraction, and it provides structured information for downstream tasks, such as knowledge graph construction, auto abstracting, and machine Q \& A. This paper focuses on the classic Event Extraction task, which consists of four subtasks: 1) Detecting event mentions from several natural language texts; 2) Determining the specific types of events; 3) Finding the arguments for each event; 4) Classifying the arguments into their roles corresponding to the event. The first two subtasks are also defined as Event Detection (ED) task, while the latter two are defined as Argument Extraction (AE) task. 
\begin{figure}
    \includegraphics[width=7.5cm]{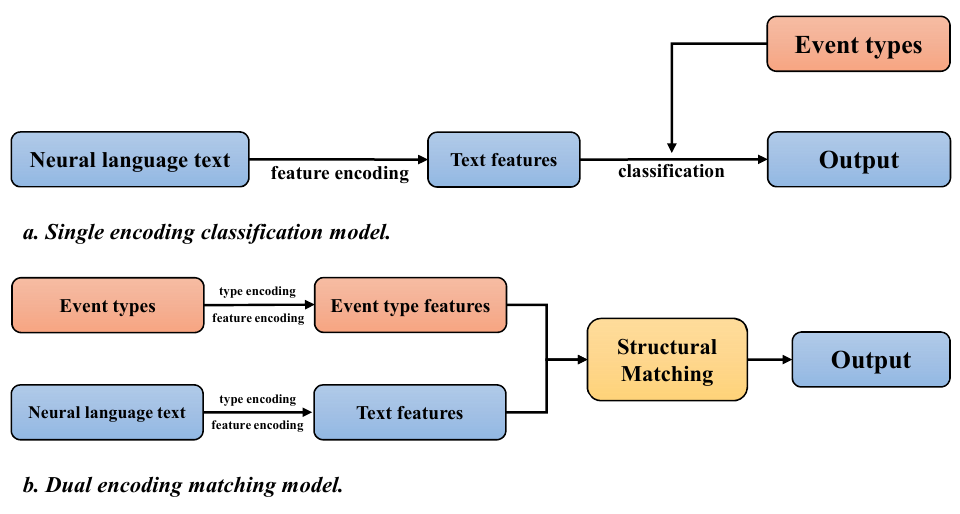}
    \caption{Single encoding and dual encoding models}
\end{figure}\label{fig:intro}
In the task of Event Extraction, some event types are specified. For example, the ACE2005 event extraction corpus is annotated with 8 types and 33 subtypes of events. Also, for each predefined event type, there is a semantic definition. For example, \textit{An \textbf{Attack} Event is defined as a violent physical act causing harm or damage.} In addition, the argument slots contained in the event type are also specified, which constitute the argument structures of the event. For example, \textbf{Attack} event contains five argument slots: \textit{Attacker, Target, Instrument, Time, and Place.} Semantic type definitions and argument structures together distinguish different event types, which are crucial in the process of human's event extraction.

In the human's way, the EE task can be abstracted as a process of matching the semantic definitions and argument structures of event types with the target text. Specifically, event mentions are detected from the text that matches the type definitions, and argument roles are extracted to fill the corresponding argument slots. The previous methods mostly focused on the feature encoding of the target text, and only introduced the event types in the final classification layer, which is called the single encoding classification models (Fig \ref{fig:intro}a). These models ignored event types' semantic and structural features, and the differences between event types are only the category numbers. As a result, these models are only sensitive to the training text. Also, they fail to correctly understand the semantic and structural connection between event types and text.

This paper proposes a dual encoding matching model (Fig \ref{fig:intro}b) to simultaneously encode event type definitions and semantic features of the target text, and structural matching is then performed to get the output. In detail, Semantic Type Embedding (STE) is first proposed to encode the semantic features of event types, entity types, and argument slot types. Then, Dynamic Structure Encoder (DSE) is used to simulate the structural matching between event types and target text. Besides, Joint Structural Semantic Matching (JSSM) model (\ref{fig:model}), a joint event extraction model based on a bidirectional attention layer is also built. JSSM achieves a significant performance improvement on the ACE2005 dataset, verifying the importance of introducing structural semantic matching.

In summary, the main contributions of this paper are as follows:
\begin{itemize}
    \item Semantic Type Embedding (STE) module is proposed to encode the semantic features of event types, entity types, and argument slot types. Utilizing these features, the Dynamic Structure Encoder (DSE) module is used to simulate the structural matching between event types and target text.
    \item Based on a bidirectional attention layer, a Joint Structural Semantic Matching (JSSM) model is built to perform event detection and argument extraction tasks jointly.
    \item Experiments on the ACE2005 dataset show that our method achieves significant performance improvement. Further analysis of the experimental results indicates the effectiveness of each module.
\end{itemize}

\section{Related Work}
The Event Extraction (EE) task can be divided into two parts: Event Detection (ED) and Argument Extraction (AE). These two parts can be solved separately or jointly.

To solve the ED task, classic methods made a hypothesis that event triggers can represent event mentions and types. \citealp{nguyen2015event} regards ED as a token-level sequence labeling problem, and uses a CNN-based classification model. Then, new models are introduced, such as Hybrid Neural Networks \cite{feng2018language}, Graph convolution networks \cite{nguyen2018graph}, Hierarchical Multi-Aspect Attention \cite{mehta2019event} and so on. Since the data in ACE2005 dataset is small in scale, researchers have introduced knowledge enhancement methods, such as FrameNet \cite{liu2016leveraging}, multilingual attention \cite{liu2018event}, and open-domain Knowledge \cite{tong2020improving}.

To solve the AE task, some models regarded it as a downstream task of ED, and these models are called the Pipeline models \cite{chen2015event,yang2019exploring}. In contrast, the Joint models \cite{nguyen2016joint, yang2016joint, sha2018jointly} take the AE and ED as interactional tasks. Pipeline models exhibit better interpretability and higher precision, but the recall rate is often lower, and the overall effect is slightly worse due to error propagation. Meanwhile, joint models often achieve better final performance owing to the information interaction between tasks.

There is a bottleneck in methods that detect event mentions and types using only event trigger words. \citealp{liu2017exploiting} exploits argument information to improve ED task, while document level information \cite{duan2017exploiting}, word sense \cite{liu2018similar}, pretrained language models \cite{yang2019exploring}, enhanced local features \cite{kan2020event}, or other features are used in the AE task. \citealp{liu2019event} also investigates the possibility of event detection without using triggers. \citealp{du2020event} and \citealp{liu2020event} treat EE as a machine reading comprehension (MRC) problem and use prior knowledge of reading comprehension to improve the model performance. 

However, previous approaches ignore the semantic and structural features of event types. This paper is the first work that explores the semantic features of event types and structurally matches them with target text to the best of our knowledge.

\section{Methods}
\subsection{Semantic Type Embedding (STE)}\label{sec:3.1}
The event extraction task aims to find the predefined event types from the text. However, if the event types are only introduced as category labels or randomly initialized vectors, event definitions' semantic features are completely lost.

Inspired by the MRC-based EE approaches \cite{du2020event, liu2020event}, the questions are replaced with event type definitions. For example, the event type \textbf{Be-Born} can be defined as \textit{"Be-Born Event occurs whenever a PERSON Entity is given birth to."}. Such a question includes semantic information that leads us to pay more attention to a person or some word standing for time.

Formally, Semantic Type Embedding (STE) module is proposed to contain two parts: static STE and dynamic STE. Given the input sentence $S=[s_{1}, s_{2}, ..., s_{n}]$ and the question $Q=[q_{1}, q_{2}, ..., q_{m}]$, a pre-trained language model, i.e., BERT (\citealp{devlin:2019}) is used as the encoder.

\subsubsection{Static STE}
First, the definition of each event type is tokenlized as the static question:
\begin{equation}
\small
    Q_{static}=[[CLS], q_{1}, q_{2}, ... q_{m}, [SEP]].
\end{equation}

After feeding these tokens into a BERT encoder, the [CLS] token's embedding is used as the static type embedding. Repeating this operation, the $ste_{static}$ for each event type can be obtained:
\begin{equation}
\small
    ste_{static}=BERT(Q_{static})_{0}
\end{equation}
 A lookup table is built, where each type corresponds to a static embedding. It is worth noting that the static STE can be generated not only for event types but also for entity types and argument slot types, because semantic definitions also exist for their types.

Static STE presents a fixed feature representation for each type. Furthermore, it is desired that the type features can be fine-tuned as the target text changes. In this case, a dynamic STE is proposed.

\subsubsection{Dynamic STE}
When definitions and target sentences are encoded together, the semantic connections between type definitions and words can be also encoded. The dynamic STE concatenates definitions of event types and target sentences, before it gets through the encoder. For example, considering a event type definition and a sentence, the input is:
\begin{equation}
\small
\begin{aligned}
     Q_{dynamic}=[ &[CLS], q_{1}, q_{2} ... q_{m}, \\
     &[SEP], w_{1}, w_{2} ... w_{n}, [SEP]].        
\end{aligned}
\end{equation}
Then, the same encoder is used:
\begin{equation}
\small
    ste_{dynamic}=BERT(Q_{dynamic})_{0}
\end{equation}
Same as the static method, each sentence is combined with definitions of every event type, and the [CLS] token's embedding is used as the dynamic STE for each type.
Finally, it is mixed with the static method with a mixing ratio $\alpha$:
\begin{equation}
\small
    ste=\alpha*ste_{static} \oplus (1-\alpha)*ste_{dynamic}
    \label{fuc:ste}
\end{equation}
In this paper, the static STE is used for entity type ($ste^{entity}$) and argument slot type ($ste^{slot}$), while the dynamic STE is used for event type ($ste^{event}$).
\begin{figure}
    \includegraphics[width=7.5cm]{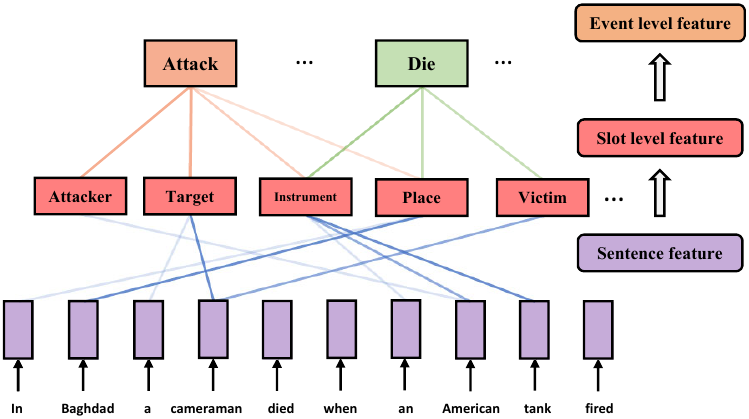}
    \caption{The structure of Dynamic Structure Encoder}
    \label{fig:DSE}
\end{figure}

\begin{figure*}[ht]
    \centering
    \includegraphics[width=16cm]{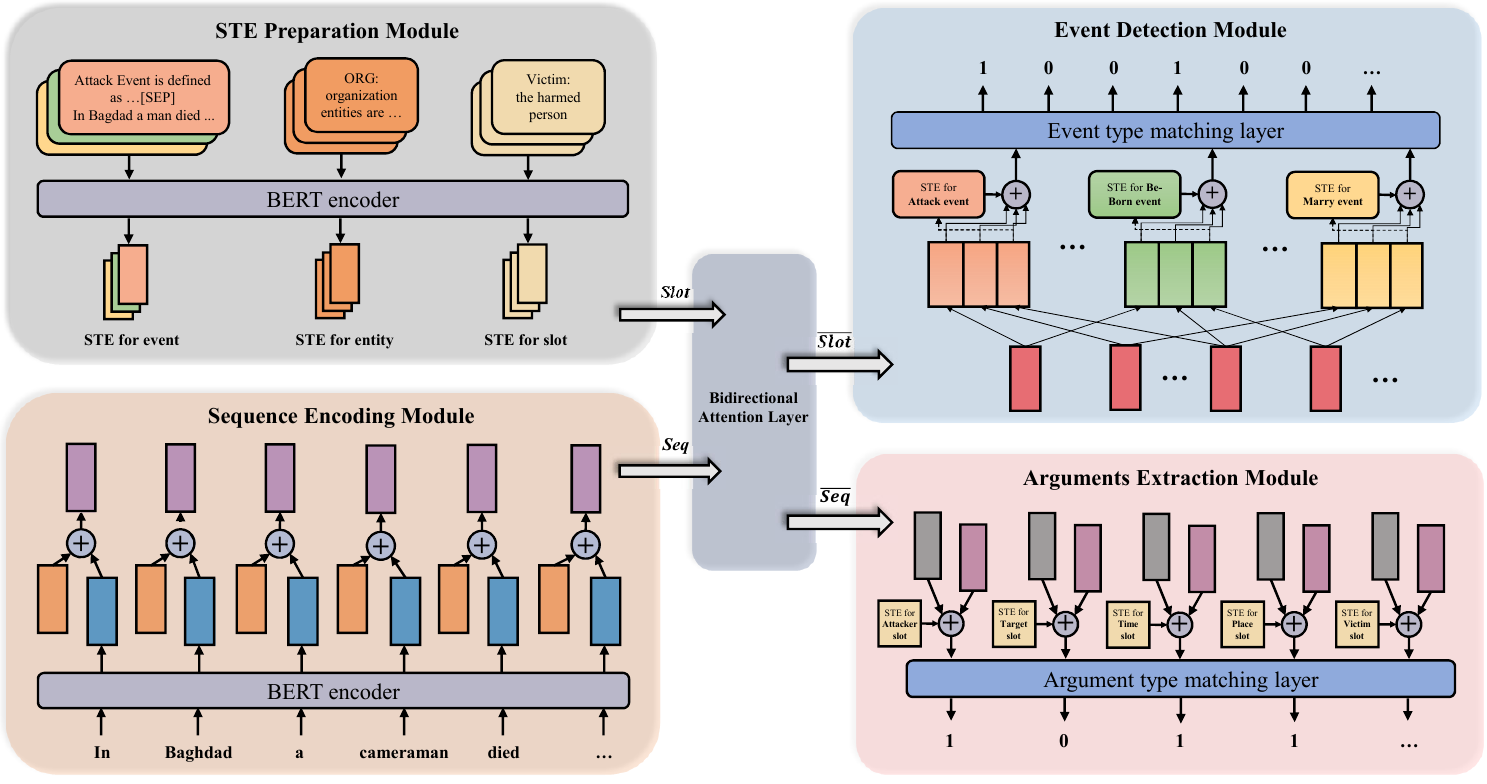}
    \caption{The overall architecture of JCMEE. ( DSE is used in the Event Detection Module)}
    \label{fig:model}
\end{figure*}
\subsection{Dynamic Structure Encoder (DSE)}\label{sec: DSE}
When humans discover events and find event roles from the text, they often use the target text to match the structure of known event types. Argument slots can be used to represent the structure of event types, because different event types have different numbers and types of argument slots. Therefore, the matching process can be abstracted as filling the argument slots of corresponding event types with words in the text.

According to the hierarchical structure of events, this paper proposes a Dynamic Structure Encoder (DSE), which has three levels of features: sentence-level feature, slot-level feature, and event-level feature. Every two levels are dynamically connected through an attention-based adding (shown in Fig \ref{fig:DSE}). 

On the Sentence-Slot connection, sentence is used to match argument slots. Taking Fig \ref{fig:DSE} as an example, after training, the slot \textbf{Target} tends to extract the features of the word "cameraman", and the slot \textbf{Instrument} tends to extract the features of the words "American" and "tank". Then, the argument slots are used to match event types based on the Slot-Event connection, and corresponding slots of each event type are only used. For example, \textbf{Attacker}, \textbf{Target}, \textbf{Instrument}, etc are used to form the event-level feature of the \textbf{Attack} event, and \textbf{Victim}, \textbf{Place}, \textbf{Instrument}, etc are used to form the event-level feature of the \textbf{Die} event.

Specifically, the input sentence can be formed as : $Sent = [w_0, w_1, ..., w_n]$, where $n$ represents the sequence length. After feature encoding, The sequence feature is $Seq = [x_{1}, x_{2}, ..., x_{n}]$.

For slot-level feature, an attention-based adding is used to make the slot aggregate the sequence features:
\begin{equation}
\small
    Slot_i = \sum_{j=0}^{n}{attention(ste^{slot}_i, x_j)\times x_j}
\end{equation}
Where $i\in[0, S]$, because there are $S$ different slot types and an \textbf{None} type. The same method in \ref{sec:3.1} is employed to obtain the static STE output $ste^{slot}_i$ for each slot, and a bidirectional attention layer is used to obtain attention scores (see \ref{sec:bidirectional} for details).

For event-level feature, another attention-based adding is performed to encode the event features:
\begin{equation}
\small
    Event_k = \sum_{i=0}^{S}{attention(ste^{event}_k, Slot_i)\times Slot_i}
\end{equation}

Where $k\in[0, E]$, because there are $E$ different event types and an \textbf{None} type. A simple $Cosine$ similarity is used to obtain the attention values. 

After encoded by the DSE, each input sentence produces $E$ different event-level features for every event types, containing lexical, semantic, and structural information. In section \ref{sec:4.4}, these event-level features and event type embeddings are used to perform event detection.

\section{Model}\label{sec:model}
A joint model called Joint Structural Semantic Matching (JSSM) is proposed in this paper for EE and AE tasks. The overall architecture of JSSM is shown in Fig \ref{fig:model}. JSSM first uses an STE Preparation Module to prepare STE for event, entity, and slot types. Then, the input sentences are encoded by a Sequence Encoding Module. After sequence feature and $Ste^{slot}$ are sent to the bidirectional attention layer, sequence-aware slot features (red boxes in Fig \ref{fig:model}) and sot-aware sequence features (grey boxes in Fig \ref{fig:model}) can be obtained. The slot features and sequence features are passed to the Event Detection Module and the Argument Extraction Module to obtain final outputs, respectively. 

\subsection{STE Preparation Module}
\label{sec:4.1}
Three STE modules are needed in this model, namely $Ste^{entity}$, $Ste^{slot}$, and $Ste^{event}$. The first two are static STEs, while $Ste^{event}$ is a dynamic STE. Taking $Ste^{slot}$ as an example:
\begin{equation}
\small
    ste^{slot}_i = BERT(question_i^{slot})_0, i\in[0,S]
\end{equation}
There are $S$ types of slots, and $question_i^{slot}$ stands for the definition sentence for the $i$th type of slot. $ste^{slot}_i\in R^{d}$, where $d$ follows the dimension of BERT encoder's output. $Slot = [ste^{slot}_i]$ is one of the two inputs of the bidirectional attention layer.

\subsection{Sequence Encoding Module}
\label{sec:4.2}
Given an input sentence $Sent = [s_i]$ with $n$ tokens, the same BERT encoder is used in STE preparation. After BERT encoding, the sequence embeddings $E = [e_{i}]$ can be obtained.
\begin{equation}
\small
    E = BERT(Sent), E\in R^{n\times d}
\end{equation}
At the same time, each token $s_i$ corresponds to a certain entity type, and $ste^{entity}_i$ is used to represent the STE of the entity. Because the same BERT encoder is used, $ste^{entity}_i\in R^d$, and they can be added together:
\begin{equation}
\small
    seq_i = e_i + ste^{entity}_i, i\in [1, n]
\end{equation}
Then, the sequence encoding is obtained: $Seq = [seq_i]$, $Seq\in R^{n\times d}$, which is the second input of the bidirectional attention layer.

\subsection{Bidirectional Attention Layer}
\label{sec:bidirectional}

More attention should be paid to the structural features in performing the event detection task, and slot features can be used (section \ref{sec: DSE}). At the same time, in performing the argument extraction task, the sequence features are more important. To jointly solve these two tasks, a bidirectional attention layer is proposed, and the two directions are called \textbf{seq2slot} and \textbf{slot2seq}. Unlike the previous joint model that directly adds the losses of different tasks, our model treats event classification and argument extraction as two mirror tasks, and the tasks enhance each other's results through the attention layer.

Assuming the STE for argument slots is $Slot = [ste_{i}^{slot}], i\in[0, S]$, and the sequence encoding $Seq = [seq_{j}], j\in[1, n]$. A method similar to the self-attention mechanism (\citealp{Vaswani:2017}) is used.
\begin{equation}
\small
    \begin{aligned}
        &Attention(Q,K,V)=softmax(\frac{QK^{T}}{\sqrt{d_{k}}})V\\
    \end{aligned}
\end{equation}
Three weight matrices are set for each way. So, there are six matrices in total, namely, $W_{V}^{slot}, W_{Q}^{slot}, W_{K}^{slot}$ and $W_{V}^{seq}, W_{Q}^{seq}, W_{K}^{seq}$.

\begin{figure}[tp]
    \centering
    \includegraphics[width=7.5cm]{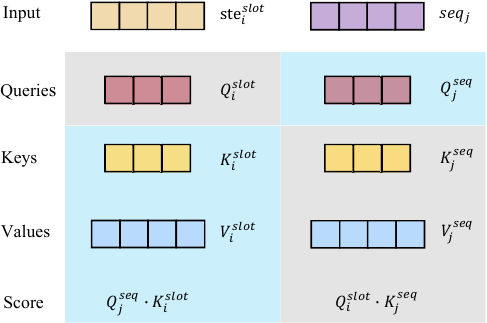}
    \caption{An example for bidirectional attention}
    \label{fig:bidirectional}
\end{figure}

\textbf{seq2slot}: In the first direction, it is desired to fill the argument slots with sequence features. So, $Slot$ is used to generate the queries, and $Seq$ is used to generate the keys and values (grey part in Fig \ref{fig:bidirectional}):
\begin{equation}
\small
    \begin{aligned}
        Q^{slot} =& Slot\cdot W_{Q}^{slot}\\
        K^{seq} =& Seq\cdot W_{K}^{seq}\\
        V^{seq}=& Seq\cdot W_{V}^{seq}\\
    \end{aligned}   
\end{equation}
Then, the sequence-aware slot features $\overline{Slot}$ ( slot-level features in DSE) can be obtained, which are red boxes in Fig \ref{fig:model}: 
\begin{equation}
\small
    \overline{Slot} = Attention( Q^{slot}, K^{seq}, V^{seq}) 
\end{equation}

\textbf{seq2slot}: In the second direction, argument slot features are used to enrich the sequence features, which is beneficial to the Argument Detection task. So, $Q^{seq}$, $K^{slot}$, and $V^{slot}$ are calculated (lake blue part in Fig \ref{fig:bidirectional}). Then, the slot-aware sequence features are obtained:
\begin{equation}
\small
    \overline{Seq} = Attention( Q^{seq}, K^{slot}, V^{slot}) 
\end{equation}
$\overline{Slot}$ and $\overline{Seq}$ are sent to the Event Classification Module and the Argument Extraction Module, respectively.

\begin{table*}[ht]
\begin{tabular}{l|ccc|ccc|ccc}
\hline
\multirow{2}{*}{Methods}& \multicolumn{3}{c|}{Event ID \& CLS}& \multicolumn{3}{c|}{Argument ID}&
\multicolumn{3}{c}{Argument CLS}\\
\cline{2-10} 
 & P & R & F1 & P & R & F1  & P & R & F1\\ 
\hline
DMCNN(\citealp{chen2015event})       &75.6 &63.6 &69.1 &68.8 &51.9 &59.1 &62.2 &46.9 &53.5 \\ 
RBPB(\citealp{sha2016rbpb})        &70.3 &67.5 &68.9 &63.2 &59.4 &61.2 &54.1 &53.5 &53.8 \\ 
\hline
JRNN (\citealp{nguyen2016joint})        &66.0 &73.0 &69.3 &61.4 &64.2 &62.8 &54.2 &56.7& 55.4 \\
Joint3EE (\citealp{nguyen2019one})      &68.0 &71.8 &69.8 &59.9 &59.8 &59.9 &52.1 &52.1& 52.1 \\
DyGIE++ (\citealp{DYGIE++})          &- &- &73.6 &- &- &55.4 &- &- &52.5 \\
\hline   
BERT-QA (\citealp{du2020event})   &  71.1& 73 &72.4&58.9 &52.1& 55.3&\textbf{56.8} &50.2& 53.3  \\
MQAEE (\citealp{Li:2020})  &  -& - &73.8&-&-&56.7&-&-&55.0  \\
\hline 
\textbf{JSSM} & \textbf{76.8} & \textbf{78.1} & \textbf{77.4} & \textbf{69.2} & \textbf{68.4} & \textbf{68.8} & 55.6 & \textbf{57.9} & \textbf{56.7}
\end{tabular}
\caption{Results on the test set. (The baseline results are directly cited from corresponding papers, and the JSSM results are the average of ten experiments with the same setting)}
\label{table:2}
\end{table*}

\subsection{Event Detection Module}\label{sec:4.4}
For each sentence input, DSE (section \ref{sec: DSE}) is used to obtain the event-level features. The input then goes through an event type matching layer to get the results.

The prepared $Ste^{event}$ is used as the type features for every event type, and the event definitions in the ACE English event guidelines are used. For each event type $e_{k}$, its type embedding $ste^{event}_k$ is used to match the input $\overline{Slot}$:

Firstly, since the event type and corresponding slot are already predefined, an event to slot $Mask$ with a hard attention map is used to mask the unrelated slot. Then, a $Cosine$ attention adding is used to get event-level features $Event_k$:
\begin{equation}
\small
    Event_k = [ste^{event}_k * (Mask * \overline{Slot})]\cdot\overline{Slot}
\end{equation}
Finally, a two-layer affine network is used to perform the matching process:
\begin{equation}
\small
\begin{aligned}
        o^{tmp}_k = \sigma[(Event_k+&ste^{event}_k) * W^{evt}_1 + b^{evt}_1] \\
        o^{evt}_k = \sigma(o^{tmp}_k *& W^{evt}_2 + b^{evt}_2)
\end{aligned}
\end{equation}
Where $W_{evt}^1, W_{evt}^2$ and $b^{evt}_1, b^{evt}_2$ are parameters,
$\sigma$ is an GELU activation function (\cite{DBLP:journals/corr/HendrycksG16}). Because each sentence may contain more than one types of event, $E$ affine networks are defined for every event types. Assuming a sentence contains event types A and C, the output should be [1, 0, 1].

The output $O^{evt}=[ o^{evt}_i ], i\in[1,E]$ and the golden label $Y^{evt} = [y^{evt}_i], i\in[1,E]$ is used to calculate the MSE loss:
\begin{equation}
\small
    L_{evt}=\frac{1}{E}\sum_{i=1}^{E}(y^{evt}_{i} - o^{evt}_i)^2
\end{equation}


\subsection{Argument Extraction Module}
$\overline{Seq}$ is used as main features for argument extraction, which contains the features of argument slots by the \textbf{slot2seq} attention. However, $\overline{Seq}$ does not contain the lexical features for each word, so the original sequence features are also added (purple boxes in Fig \ref{fig:model}).
\begin{equation}
\small
     Arg_i = \overline{Seq}_i + seq_i
\end{equation}
Since each word may have multiple argument labels, similar affine networks in section \ref{sec:4.4} are used. 
\begin{equation}
\small
\begin{aligned}
        o^{tmp}_k = \sigma[(Arg_k+&ste^{slot}_k) * W^{arg}_1 + b^{arg}_1] \\
        o^{arg}_k = \sigma(o^{tmp}_k *& W^{arg}_2 + b^{arg}_2)
\end{aligned}
\end{equation}
Assuming the golden labels are $Y^{arg}=[y^{arg}_i],i\in [1,S]$, the loss for arguments extraction can be obtained:
\begin{equation}
\small
     L_{arg}=\frac{1}{S}\sum_{i=1}^{S}(y^{arg}_{i} - o^{arg}_i)^2
\end{equation}
Combining $L_{evt}$ and $L_{arg}$, the overall loss function can be obtained.
\begin{equation}
\small
    L = \lambda L_{evt}+(1-\lambda L_{arg})
\label{fuc:loss}
\end{equation}
\section{Experiments}

\begin{table*}[]
\centering
\begin{tabular}{l|ccc|ccc|ccc}
\hline
\multirow{2}{*}{Methods}& \multicolumn{3}{c|}{Event ID \& CLS}& \multicolumn{3}{c|}{Argument ID}&
\multicolumn{3}{c}{Argument CLS}\\
\cline{2-10} 
      & P & R & F1 & P & R & F1  & P & R & F1\\ \hline
JSSM & 76.8 & 78.1 & 77.4 & 69.2 & 68.4 & 68.8 & 55.6 & 57.9 & 56.7\\ 
\quad$-DSE$ & 74.6 & 77.4 & 76.0 & 63.0 & 65.7 & 64.3 & 50.5 &  52.8 & 51.6 \\
\quad$-Ste_{dynamic}$ & 75.8 & 77.9 & 76.8 & 69.6 & 67.1 & 68.3 & 56.2 & 56.7 & 56.4\\
\quad$-Ste^{entity}$ & 72.9 & 76.1 & 74.5 & 56.1 & 67.9 & 61.4 & 45.0 & 54.4 & 49.3 \\
\quad $random\; Ste^{entity}$ & 74.7 & 76.7 & 75.7 & 63.3 & 63.7 &  63.5 & 52.1 & 52.5 & 52.3  \\
\quad $random\; Ste^{slot}$ &  75.6 & 74.6 & 75.1 & 63.7 & 64.4 & 64.0 & 51.4 & 52.0 & 51.7  \\
\quad $random\; Ste^{event}$ & 79.1  & 49.5 & 61.0  & 64.2 & 59.6 & 61.8 & 53.9 & 50.1 & 51.9\\
\end{tabular}
\caption{Ablation study. (Each result is the average of ten experiments with the same setting)}
\label{table:3}
\end{table*}

\subsection{Dataset and Settings}
\textbf{Dataset:} The proposed method is evaluated on the  ACE2005 (\citealp{ACE2005}) English event extraction benchmark. Annotated from 599 documents, the benchmark contains 33 types of event. Following the previous splitting and processing settings (\citealp{nguyen:2016}, \citealp{Zhang:2019}, \citealp{liu2019event}), 529 documents are used for train data, 40 documents for test and 30 for dev.

\textbf{Environment and Hyper parameters:} Experiments are performed on a single Nvidia gpu rtx2080ti. Using BERT-base as the pretrained language model, all embedding and features' dimensions are unified to 768. The batch size is 16. The learning rate is set to 1.5e-5, and the dropout rate is 0.6. The mixing ratio $\alpha$ in dynamic STE (Eq: \ref{fuc:ste}) and $\lambda$ in loss function (Eq: \ref{fuc:loss}) are set to 0.5 and 0.3, respectively. Each epoch takes about 20 minutes. 

\textbf{Evaluation Metrics:} For the event detection task, instead of locating the trigger, whether the event types of a sentence match the golden labels is directly judged. For the argument extraction task, an argument is correctly identified if its span matches a golden annotated argument. For the argument classification task, an argument is correctly classified if its span and role both match the gold annotation. The \textbf{P}recision, \textbf{R}ecall and macro \textbf{F1} scores are calculated for each task. 

\subsection{Baselines}
The following models are used as baselines.

\textbf{Two pipeline models:} 1) DMCNN(\citealp{chen2015event}), which extracts sentence-level features by adopting a dynamic multi-pooling CNN model. 2) RBPB(\citealp{sha2016rbpb}), which proposed a regularization-based method to utilize the pattern of arguments.

\textbf{Three joint models:} 3) JRNN (\citealp{nguyen2016joint}), which uses an RNN-based joint model. 4) Joint3EE (\citealp{Joint3EE}) and which jointly models entities and events based on shared Bi-GRU hidden representation. 5)  DyGIE++(\citealp{DYGIE++}), which jointly extracts entity, relation, and event.

\textbf{Two MRC-based models:} 5) BERT-QA(\citealp{du2020event}), which formulates event extraction as a question answering problem; 6) MQAEE (\citealp{Li:2020}), which performs event extraction based on a multi-turn QA framework.

\subsection{Overall Performance}
The overall performance is listed in table \ref{table:2}. The \textbf{Event ID\& CLS} column compares precision, recall and F1 scores of the event detection task. The \textbf{Argument ID} and \textbf{Argument CLS} columns compare performance of the argument extraction task.

It can be observed that: 1) By using structural semantic matching, the JSSM model's overall performance outperforms prior works on all tasks. 2) The recall scores of JSSM significantly exceed all baselines, while the precision score of the arguments classification task is relatively weak. 3) The drop of F1 from argument identification to classification is 12.1 \%, which is much larger than prior models. It is speculated that our joint model loses certain prior information while avoiding error propagation.

\subsection{Ablation Study}\label{ablation:ablation study}
To clarify each modules' performance contribution to the model, some ablation studies are conducted, and the results are listed in table \ref{table:3}. In these studies, parts of our model are deactivated separately. In table \ref{table:3}, "$-$" means that the module is abandoned, while "$random$" means that randomly initialized embedding is used to replace the STE module.

After the $DSE$ module is removed, the F1 score of Event ID \& CLS drops by 1.4\%. Furthermore, the F1 score of Argument ID and Argument CLS drops by 4.5\% and 5.1\%, respectively. Though the DSE module does not directly act on the Argument ID \& CLS tasks, these tasks' performance exhibits a significant decrease, which reflects our model connects the structural features of event types and argument slots through a joint training process.

When the $Ste_{dynamic}$ is removed, the F1 scores drop slightly on all tasks. It is worth mentioning that only using static STE reduces noise disturbance and has higher precision scores in argument extraction tasks.

The experiment "$-Ste^{entity}$" shows that after entity labels are abandoned, the model suffers from a sharp performance decrease, 7.4\% in both argument identification and classification. These results indicate that entity features are one of the most useful information in sequence classification tasks.

In the three $random$ experiments, the semantic features of event, entity, and argument slot are completely abandoned. The performance on each task decreases apparently, indicating the indispensability of semantic features. Furthermore, when the $Ste^{event}$ is randomly initialized, the structural semantic matching between sentences and event types also fails, causing a devastating performance drop in Event ID \& CLS tasks.

\begin{figure*}[th]
\centering
    \includegraphics[width=16cm]{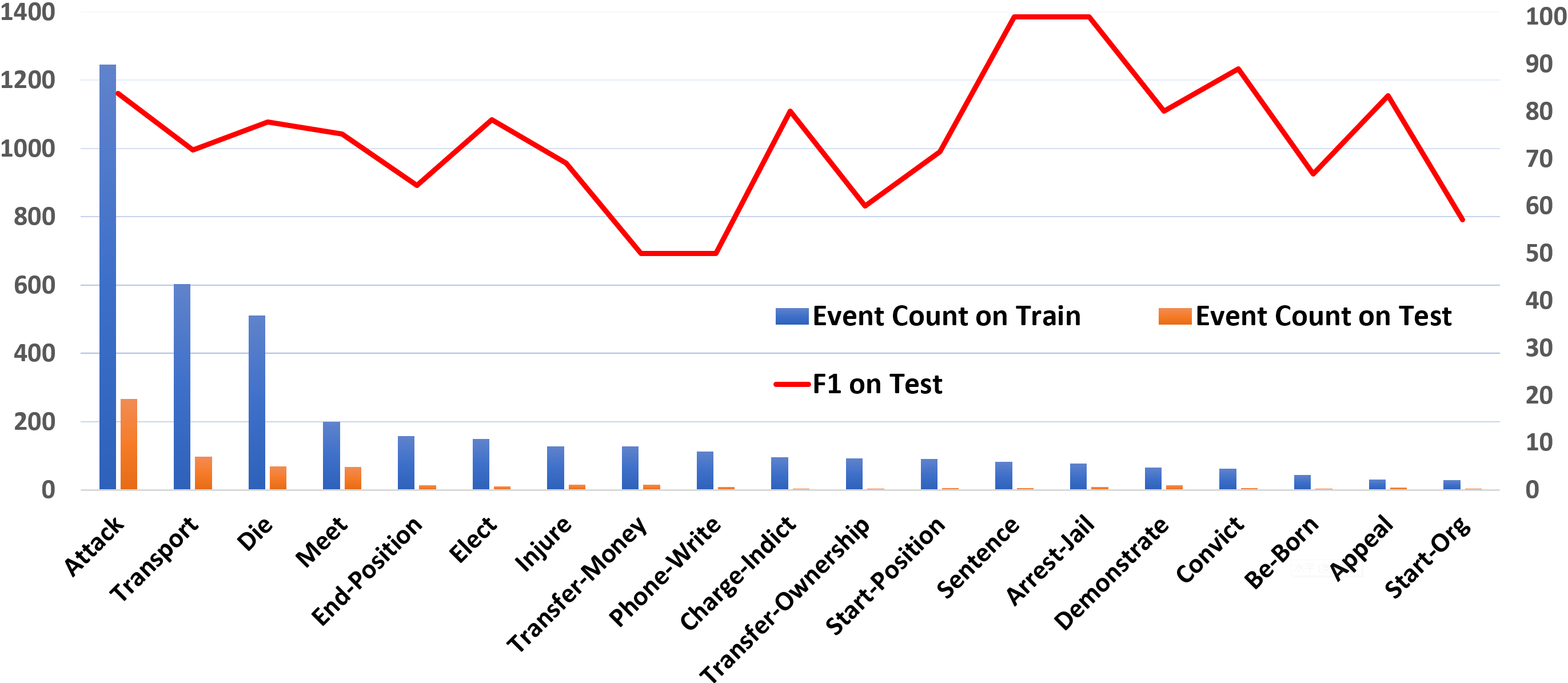}
    \caption{Event distribution analysis. (Only include event types that occur $\geq$2 times in the test data)}
    \label{fig:count}
\end{figure*}

\begin{table*}[]
\centering
\begin{tabular}{c|ccc|ccc|ccc}
    \hline
    \multirow{2}{*}{Question Strategy} & \multicolumn{3}{c|}{Event ID \& CLS}  & \multicolumn{3}{c|}{Argument ID} & \multicolumn{3}{c}{Argument CLS} \\ \cline{2-10} 
    & \multicolumn{1}{c}{P} & \multicolumn{1}{c}{R} & F1   & \multicolumn{1}{c}{P} & \multicolumn{1}{c}{R} & F1   & \multicolumn{1}{c}{P} & \multicolumn{1}{c}{R} & F1   \\
    \hline
    Single event name & 73.9 & \textbf{78.8} & 76.3  & 64.1 & \textbf{74.7} & \textbf{69.0}  & 52.2 & \textbf{60.7} & 56.1 \\ 
    Top trigger words & 75.9 & 74.7 & 75.3  & 61.7 & 62.0 & 61.8  & 47.3 & 47.5 & 47.4 \\
    Guideline definition  & \textbf{76.8} & 78.1 & \textbf{77.4} & \textbf{69.2} & 68.4 & 68.8 & \textbf{55.6} & 57.9 & \textbf{56.7} \\
\end{tabular}
\caption{Semantic material selection for $Ste^{event}$}
\label{table:5}
\end{table*}
\subsection{Semantic Material Selection}
The semantic material used to generate questions for the STE module determines the semantic information of type embeddings. To explore the influence of different question strategies on the results, three templates focusing on the event names, triggers, and definitions are designed respectively:

\textbf{Single event name}: A single event name is used as the question, and its BERT embedding is used as the STE for event.

\textbf{Top trigger words}: For each event type, the top five most frequently occurring trigger words are chosen, and they are concatenated as the question.

\textbf{Guideline definition (used in JSSM)}: Each event type has a definition in the ACE English event guidelines. The definition is slightly modified to contain more structural information. Then, they are taken as the questions.

The results are shown in Table \ref{table:5}. It can be seen that: 1) The Guideline definition material exhibits the best effect. Because it describes the event from all aspects,  contributing to sufficient semantic information. 2) Single event name strategy achieves better performances in the Argument ID task. Owing to BERT's large-scale training corpus, the model can also generate event-related semantic information from event names. 3) Trigger words are mostly verbs, but these verbs have different meanings depending on the context, and they do not merely represent a specific event, leading to the worst results.

\subsection{Event Imbalance Analysis}
The ACE2005 dataset has a serious data imbalance problem (\cite{liu2016leveraging,tong2020improving}). Only the number of \textbf{Attack}, \textbf{Transport} and \textbf{Die} events accounted for 51.5\% of all events in the training set. Therefore, models that only use text features have poor results on rare event types.

Our model makes full use of the event type's semantic features, making it robust to data imbalance. It can be seen from Fig \ref{fig:count} that there is no significant difference in the performance from dense to rare event types.

\section{Conclusion}
In this paper, we regard the event extraction task as a structural semantic matching process between the event types and the target text. The STE module and the DSE module are proposed, and a joint extraction model named JSSM is built based on a bidirectional attention layer, which performs well on the ACE2005 benchmark. Each module's effectiveness is verified through the ablation study, and the performance of different question strategies for STE is compared. Also, our model is showed extremely robust on the imbalanced data.

$\left[
\begin{array}{c}
     \textbf{1} \\
     \textbf{0} \\
     \textbf{1} \\
     \textbf{...} \\
     \textbf{0}\\
     \textbf{0}
\end{array}
\right]$


\bibliographystyle{acl_natbib}
\bibliography{acl2021}

\end{document}